\documentclass{article}

\usepackage[english]{babel}

\usepackage[letterpaper,top=2cm,bottom=2cm,left=3cm,right=3cm,marginparwidth=1.75cm]{geometry}

\usepackage[numbers]{natbib}
\usepackage{algorithm}
\usepackage[noend]{algpseudocode}
\usepackage{float}
\usepackage{graphicx}
\algrenewcommand\algorithmicrequire{\textbf{Input:}}
\algrenewcommand\algorithmicensure{\textbf{Output:}}
\usepackage{amsmath} 

\usepackage{hyperref}
\usepackage{booktabs}
\usepackage{float}

\usepackage{booktabs}

\title{Ensemble Elastic DQN: A Step Dependent Ensemble Approach for Reducing Overestimation in Deep Value-Based Reinforcement Learning}
\author{Adrian Ly, Richard Dazeley, Peter Vamplew, Francisco Cruz and Sunil Aryal}

\begin{document}
\maketitle

\begin{abstract}
Deep Q-Networks (DQN) can suffer from overestimation bias because bootstrapped targets use a maximisation operation over noisy value estimates. Ensemble-based methods and multi-step methods have each been used to improve the stability and sample efficiency of value-based reinforcement learning, but their interaction remains less well understood. This paper introduces Ensemble Elastic DQN (EEDQN), a value-based reinforcement learning algorithm that combines adaptive elastic multi-step returns with ensemble-based target aggregation. EEDQN replaces the clustering-based state similarity test used in earlier Elastic Step DQN with a lightweight Q-value difference rule, making adaptive return construction simpler to apply in discrete control settings. The method then applies horizon dependent ensemble aggregation, one-step targets use the ensemble mean, while longer elastic returns use the ensemble minimum. This design aims to reduce extreme optimistic bootstrap estimates without making every update uniformly conservative. We evaluate EEDQN on five MinAtar environments against DQN, Double DQN, Averaged DQN, MaxMin DQN, and Elastic Step DQN. EEDQN achieves the highest final return in four of the five environments and remains competitive with conservative ensemble baselines. An aggregation rule ablation shows that the best degree of conservatism is environment dependent, suggesting that adaptive return length and ensemble aggregation interact in non-trivial ways.
\end{abstract}

\vspace{0.5em}
\noindent\textbf{Keywords:} deep reinforcement learning; DQN; overestimation bias; ensemble learning; multi-step returns
\vspace{1em}

\section{Introduction}

Overestimation bias is a persistent issue in value-based deep reinforcement learning. In DQN, bootstrapped targets use a maximisation operation over estimated action values, so noisy positive errors can be selected and propagated through subsequent updates \citep{mnih2015human, van2016deep}. This can inflate Q-value estimates, mislead action selection, and contribute to convergence toward sub-optimal policies. The issue is not limited to DQN and has also been observed in other value-based and actor-critic methods \citep{fujimoto2018addressing, nauman2024overestimation}. When combined with recursive bootstrapping, small positive errors can be amplified over training, making overestimation an important source of instability in deep reinforcement learning \citep{thrun2014issues, hessel2018rainbow, he2020wd3, van2018deep}. Inflated Q-values may also affect exploration by encouraging agents to revisit state-action regions whose values have been overestimated \citep{zhang2025beta}. As a result, several algorithmic extensions have been proposed to reduce estimation error and improve learning stability. Two particularly relevant directions are ensemble-based value estimation and multi-step bootstrapping.

Ensemble-based algorithms address the problem of overestimation in DQN through the use of multiple independent approximators to moderate Q-value estimations. Algorithms like Averaged DQN \citep{anschel2017averaged} and MaxMinDQN \citep{lan2020maxmin} aggregate the predictions of multiple parallel Q-value approximators; the combination of multiple Q-networks reduces the effect of overestimated values from any one Q-network's outliers, thereby smoothing out inflated estimates and providing more reliable targets for learning.  

On the other hand, multi-step methods such as Rainbow \citep{hessel2018rainbow} or Elastic Step DQN \citep{ly2024elastic}  bootstrap returns over a longer horizon and reduce the reliance on single-step bootstrapping, which has shown reductive effects on overestimation in DQN and  strong sample efficiency by accelerating credit assignment \citep{hernandez2019understanding, yuan2019novel, hessel2018rainbow, ly2024elastic}.

In general, research around the integration of different algorithmic extensions to improve performance is scarce. While heuristic approaches like ensemble-based methods and multi-step algorithms show a lot of promise, they have largely been explored in isolation. Both areas of research improve algorithmic performance by changing how state-action value estimates are constructed, aggregated, or bootstrapped, albeit through different mechanisms. This raises the question of whether these approaches are complementary in the context of alleviating Q-value overestimation. While the benefits are clear, the combination of the two heuristic approaches is non-trivial, as the reductive effects of the combination of both approaches can lead to underestimation bias.

Elastic Step DQN, introduced by \citep{ly2023elastic, ly2024elastic}, is an adaptive extension of multi-step DQN that adjusts the backup horizon online according to the similarity between consecutive states. It accumulates rewards over longer horizons in stable regions and shortens the horizon when the trajectory appears to enter a different region of the state space. However, the original method relies on a clustering-based similarity test, and its one-step returns remain susceptible to overestimation bias. This paper introduces Ensemble Elastic DQN (EEDQN), which replaces the clustering step with a lightweight Q-value difference rule and combines elastic return construction with horizon dependent ensemble aggregation. The goal is to reduce extreme optimistic bootstrap targets while preserving the sample efficiency benefits of adaptive multi-step learning. The contributions of this paper are as follows:

\begin{itemize}
  \item We introduce Ensemble Elastic DQN (EEDQN), a value-based reinforcement learning algorithm that combines adaptive elastic multi-step returns with ensemble-based target aggregation.
  \item We replace the clustering-based similarity test used in Elastic Step DQN with a lightweight Q-value difference rule based on ensemble target values, reducing the additional machinery required for adaptive return construction.
  \item We propose a horizon dependent aggregation rule in which one-step targets use the ensemble mean, while longer elastic returns use the ensemble minimum.
  \item We evaluate EEDQN against DQN, Double DQN (DDQN), Averaged DQN (AvgDQN), MaxMin DQN, and Elastic Step DQN (ESDQN) across the five MinAtar environments using 10 matched random seeds.
  \item We conduct an aggregation rule ablation over mean, minimum, swapped, and convex aggregation variants to analyse how ensemble conservatism interacts with adaptive elastic return length.
\end{itemize}

\section{Related work}

A central source of instability in Q-learning is the use of a maximisation operation over noisy value estimates. Double DQN addresses this issue by decoupling action selection from action evaluation, thereby reducing the tendency of DQN to propagate overly optimistic targets \citep{van2016deep}. Ensemble-based methods provide another way to control estimation error. Averaged DQN reduces variance by averaging the predictions of multiple value functions \citep{anschel2017averaged}, while MaxMin DQN uses the minimum value across an ensemble to produce more conservative bootstrap targets \citep{lan2020maxmin}. These methods show that aggregating multiple value estimates can improve stability, but they differ in how aggressively they suppress overestimated values.

Multi-step methods address a related problem from a different direction. Instead of relying only on a one-step bootstrap target, an $n$-step return accumulates rewards over a longer horizon before bootstrapping. This can improve credit assignment and sample efficiency, but the choice of $n$ is fixed before training and may not be appropriate across all regions of the state space \citep{hessel2018rainbow, hernandez2019understanding, yuan2019novel}. Elastic Step DQN extends this idea by adapting the backup horizon online according to a similarity test between successive states \citep{ly2023elastic, ly2024elastic}. However, the original Elastic Step DQN used representation clustering to decide when to stop accumulating rewards, which can become computationally expensive as the state representation and environment complexity increase.

Comparatively less work has examined how ensemble aggregation and adaptive multi-step returns should be combined. \cite{chen2022ensemble} combine ensemble and multi-step learning in a hierarchical actor-critic setting, where base learners exchange information and a global critic evaluates the base learners. This differs from the setting studied here, where the goal is to combine elastic multi-step returns with parallel Q-network aggregation in a DQN-style value-based algorithm. EEDQN is therefore positioned as a lightweight value-based approach that studies how the aggregation rule should depend on the adaptive backup horizon.

EEDQN is also related to broader work on controlling estimation bias in deep reinforcement learning. Double DQN (\cite{van2016deep}) reduces overestimation by separating action selection from action evaluation, while MaxMin DQN (\cite{lan2020maxmin}) introduces conservatism through a minimum over multiple value estimates. Multi-step targets reduce the relative influence of the bootstrap term, but do not by themselves determine whether the remaining bootstrap value should be optimistic, averaged, or conservative. The contribution of EEDQN is not the use of ensembles or multi-step returns in isolation, but the study of a horizon dependent rule that changes the aggregation operator according to the adaptive return length. This distinguishes EEDQN from methods that apply a single aggregation rule uniformly across all targets.

\section{Ensemble Elastic DQN}
\subsection{Elastic Step DQN}

Standard $n$-step DQN methods use a fixed backup horizon. This horizon must be selected before training and may not be appropriate across all parts of the state space. Elastic Step DQN addresses this limitation by adapting the backup horizon online. When consecutive transitions appear similar, rewards are accumulated over a longer trajectory before bootstrapping. When the trajectory enters a less similar region, the accumulated transition is stored and learning proceeds from that variable length return \citep{ly2023elastic, ly2024elastic}.

The original Elastic Step DQN implementation measured similarity using representations extracted from the Q-network and then applied HDBSCAN clustering to determine whether successive states belonged to the same region. This approach was effective in smaller control environments, but the clustering step introduces additional computational cost and makes the method harder to scale to larger benchmark settings. EEDQN therefore replaces the clustering-based similarity test with a lightweight Q-value difference rule.

\subsection{Scalable Elastic Step DQN}

To avoid the clustering cost of the original Elastic Step DQN, EEDQN uses changes in the ensemble target values as a lightweight proxy for transition similarity. Let $\{\hat Q_i\}_{i=1}^{N}$ denote the target networks of an ensemble of $N$ value functions, and define the ensemble mean target value vector as
\[
\bar Q^{-}(s,\cdot) = \frac{1}{N}\sum_{i=1}^{N}\hat Q_i(s,\cdot).
\]
For the current accumulation start state $s_0$ and the most recent next state $s'$, EEDQN computes the value-difference statistic
\[
z = \left\| \bar Q^{-}(s_0,\cdot) - \bar Q^{-}(s',\cdot) \right\|_1 .
\]
A historical buffer $B$ stores recent values of $z$. Before adding the current value to the buffer, EEDQN computes the adaptive threshold
\[
h = \operatorname{mean}(B) + \frac{\operatorname{std}(B)}{\sqrt{|B|}}.
\]
If $z > h$, the trajectory is treated as having entered a different value region, so the accumulated transition is stored. Otherwise, reward accumulation continues. The current $z$ is then appended to $B$ for use in later threshold estimates.

\begin{algorithm}[H]
\caption{EEDQN target calculation}
\label{alg:eedqn_target}
\begin{algorithmic}[1]
\Require Accumulated return $R^{(k)}$, next state $s'$, horizon $k$, terminal flag $\textit{done}$, target ensemble $\{\hat Q_i\}_{i=1}^{N}$, discount factor $\gamma$
\Ensure Target value $y$

\If{$\textit{done}$}
    \State \Return $R^{(k)}$
\EndIf

\State Compute the ensemble-mean target value for each action:
\[
\bar Q^{-}(s',a) = \frac{1}{N}\sum_{i=1}^{N}\hat Q_i(s',a)
\]
\State Select the greedy bootstrap action:
\[
a^{*} \gets \arg\max_a \bar Q^{-}(s',a)
\]

\If{$k = 1$}
    \State $G \gets \dfrac{1}{N}\sum_{i=1}^{N}\hat Q_i(s',a^{*})$
\Else
    \State $G \gets \min_i \hat Q_i(s',a^{*})$
\EndIf

\State \Return $R^{(k)} + \gamma^k G$
\end{algorithmic}
\end{algorithm}

\begin{algorithm}[H]
\caption{Adaptive value-difference threshold}
\label{alg:eedqn_threshold}
\begin{algorithmic}[1]
\Require Start state $s_0$, next state $s'$, target ensemble $\{\hat Q_i\}_{i=1}^{N}$, value-difference buffer $B$
\Ensure Value difference $z$ and adaptive threshold $h$

\State Compute ensemble-mean target vectors:
\[
\bar Q^{-}(s_0,\cdot)
=
\frac{1}{N}\sum_{i=1}^{N}\hat Q_i(s_0,\cdot),
\qquad
\bar Q^{-}(s',\cdot)
=
\frac{1}{N}\sum_{i=1}^{N}\hat Q_i(s',\cdot)
\]

\State Compute the value-difference statistic:
\[
z \gets \left\|\bar Q^{-}(s_0,\cdot)-\bar Q^{-}(s',\cdot)\right\|_1
\]

\State Compute the adaptive threshold:
\[
h \gets \operatorname{mean}(B) + \frac{\operatorname{std}(B)}{\sqrt{|B|}}
\]

\State \Return $z,h$
\end{algorithmic}
\end{algorithm}

\begin{algorithm}[H]
\caption{Ensemble Elastic Deep Q-Network}
\label{alg:eedqn}
\begin{algorithmic}[1]
\Require Environment, replay buffer $D$, value-difference buffer $B$, ensemble size $N$
\Ensure Trained ensemble $\{Q_i\}_{i=1}^{N}$

\State Initialise online networks $\{Q_i\}_{i=1}^{N}$
\State Initialise target networks $\{\hat Q_i\}_{i=1}^{N}$
\State Pre-fill replay buffer $D$ with random transitions
\State Pre-fill value-difference buffer $B$ with initial value differences

\For{each episode}
    \State Reset the environment
    \State Select the initial action using an $\epsilon$-greedy ensemble-mean policy
    \State Initialise accumulated return $R^{(0)} \gets 0$
    \State Initialise horizon $k \gets 0$

    \For{each environment step}
        \State Execute the current action and observe reward $r$, next state $s'$, and $\textit{done}$
        \State Update the accumulated discounted return $R^{(k)}$
        \State Compute $z$ and $h$ using Algorithm~\ref{alg:eedqn_threshold}

        \If{$z > h$ or $\textit{done}$}
            \State Store the accumulated transition in $D$
            \If{not $\textit{done}$}
                \State Restart accumulation from the current state
            \EndIf
        \EndIf

        \If{ready to update}
            \State Sample a mini-batch from $D$
            \State Compute targets using Algorithm~\ref{alg:eedqn_target}
            \State Update each ensemble member by minimising squared TD error
        \EndIf

        \If{target update step}
            \State Copy online parameters to target parameters
        \EndIf
    \EndFor
\EndFor
\end{algorithmic}
\end{algorithm}

\subsection{EEDQN training procedure}

EEDQN (Algorithm \ref{alg:eedqn}) maintains an ensemble of $N$ online Q-networks and corresponding target networks. Each ensemble member is independently initialised and is updated using the same sampled target, while action selection uses an $\epsilon$-greedy policy with respect to the ensemble-mean online value estimate. During environment interaction, EEDQN accumulates rewards from a start state $s_{\mathrm{start}}$ until the value-difference statistic between $s_{\mathrm{start}}$ and the current next state exceeds the adaptive threshold. The stored replay transition therefore has the form
\[
(s_{\mathrm{start}}, a_{\mathrm{start}}, R^{(k)}, s', k, \textit{done}),
\]
where $k$ is the number of environment rewards accumulated and
\[
R^{(k)}=\sum_{j=0}^{k-1}\gamma^{j}r_{t+j}.
\]

For non-terminal transitions, EEDQN first selects the greedy bootstrap action using the ensemble-mean target value,
\[
a^{*} = \operatorname*{arg\,max}_{a}\frac{1}{N}\sum_{i=1}^{N}\hat Q_i(s',a).
\]
It then applies a different aggregation rule depending on the accumulated horizon. For one-step transitions, the bootstrap value is the ensemble mean,
\[
G_{\mathrm{mean}}(s',a^{*})=\frac{1}{N}\sum_{i=1}^{N}\hat Q_i(s',a^{*}),
\]
whereas for multi-step transitions, the bootstrap value is the minimum over the ensemble,
\[
G_{\mathrm{min}}(s',a^{*})=\min_i \hat Q_i(s',a^{*}).
\]
The resulting target is
\[
y =
\begin{cases}
R^{(k)}, & \text{if terminal},\\
R^{(1)}+\gamma G_{\mathrm{mean}}(s',a^{*}), & \text{if } k=1,\\
R^{(k)}+\gamma^{k}G_{\mathrm{min}}(s',a^{*}), & \text{if } k>1.
\end{cases}
\]

The motivation for using the mean on one-step targets is that these targets rely heavily on bootstrapping and can become overly conservative if the minimum is used everywhere. For longer elastic returns, more reward information has already been accumulated before the bootstrap term, so a conservative ensemble minimum can be used to reduce optimistic extrapolation while having less influence over the full target. Since this design choice is not guaranteed to be optimal in every environment, Section~\ref{sec:ablation} compares it against alternative aggregation rules.

\section{Empirical evaluation}
\subsection{Experimental setup}

All algorithms were evaluated on the same five MinAtar environments, Seaquest, SpaceInvaders, Asterix, Breakout, and Freeway \citep{young2019minatar, obando2020revisiting}. MinAtar was selected because it preserves key Atari control challenges while substantially reducing the computational cost of experimentation, allowing each algorithm to be evaluated across 10 matched random seeds. Each agent was trained for one million environment steps. Episodic returns were aggregated into 100 training epochs for visualisation, so each plotted epoch corresponds to approximately 10,000 environment steps.

MushroomRL \citep{d2021mushroomrl} was used as the reinforcement learning framework. Unless otherwise specified in Table~\ref{tab:hyperparameters}, all algorithms used the same training budget, replay memory size, batch size, discount factor, learning rate, target update interval, and exploration schedule. Exploration used an $\epsilon$-greedy policy with a linear decay schedule. For ensemble-based methods, each ensemble member was independently initialised. The ensemble size was set to two for AvgDQN, MaxMinDQN, and EEDQN in order to study the smallest non-trivial ensemble while keeping the computational cost close to the non-ensemble baselines.

Final performance is reported as the mean episodic return over the last 100 training episodes for each seed, averaged across seeds. Statistical significance was assessed using a two-sided paired permutation test over seed-level final returns. For each environment and baseline comparison, the null hypothesis was that the paired performance differences between EEDQN and the baseline were exchangeable. At each permutation, the paired algorithm labels were randomly swapped within each seed and the difference in mean final return was recomputed. This non-parametric test was selected because the number of seeds is modest and because it does not assume normally distributed returns.

To diagnose extreme value inflation, we compare the maximum observed Q-value during training against the loose discounted-return upper bound $Q_{\max}=r_{\max}/(1-\gamma)$. For SpaceInvaders, Asterix, Breakout, and Freeway, $r_{\max}=1$, giving $Q_{\max}=100$ when $\gamma=0.99$. For Seaquest, $r_{\max}=10$, giving $Q_{\max}=1000$. Values above this bound indicate Q-values that exceed the maximum possible discounted return under the assumed reward limit.

\begin{figure*}
\includegraphics[width=0.95\textwidth]{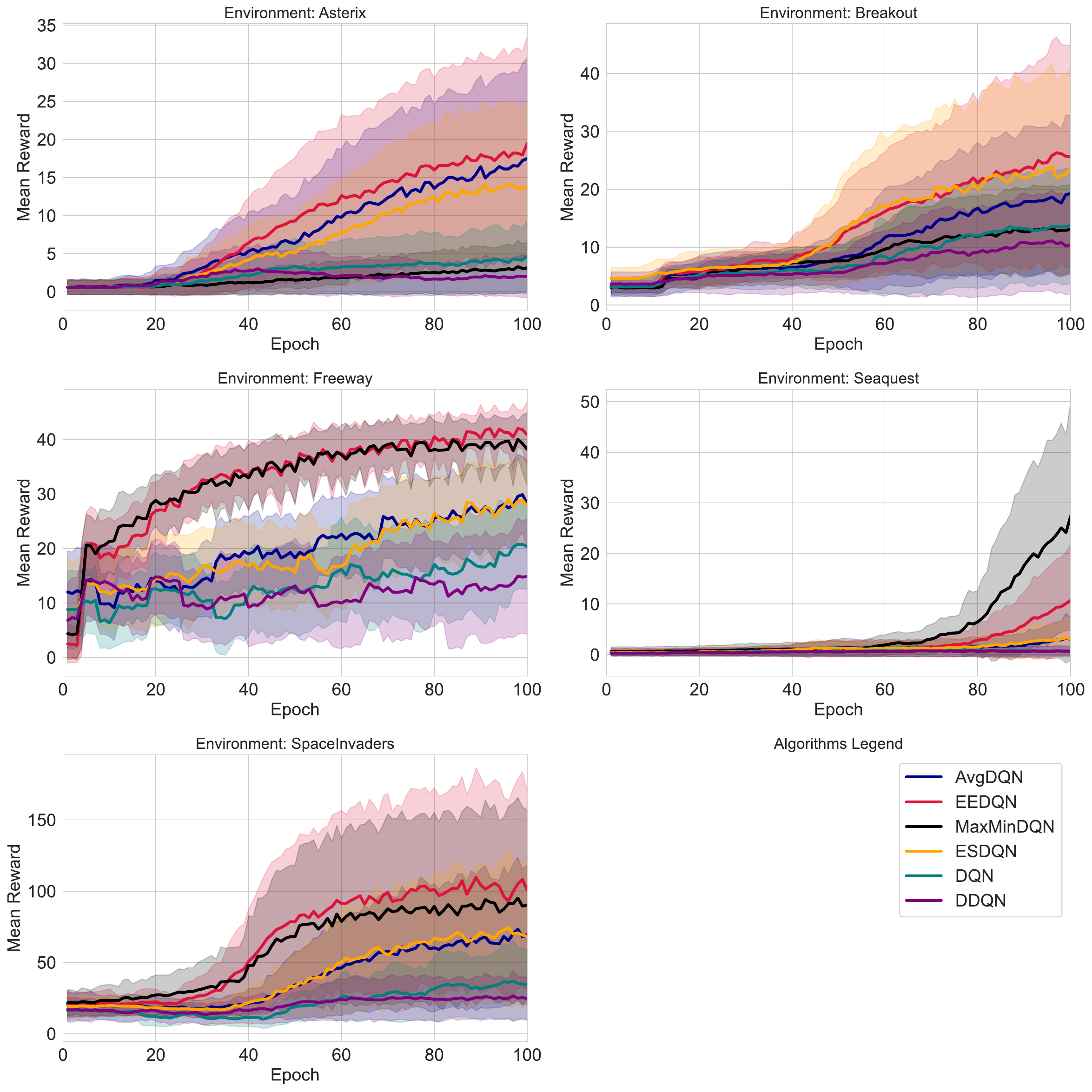}
\caption{Training performance across the five MinAtar environments over one million environment steps. The x-axis shows 100 training epochs, where each epoch corresponds to approximately 10,000 environment steps. The y-axis shows mean episodic reward. Each curve represents the mean across 10 seeds, and the shaded region indicates the 95 percent confidence interval around the mean.}
\label{fig1:average_minatar}
\end{figure*}

\begin{figure*}
\includegraphics[width=0.95\textwidth]{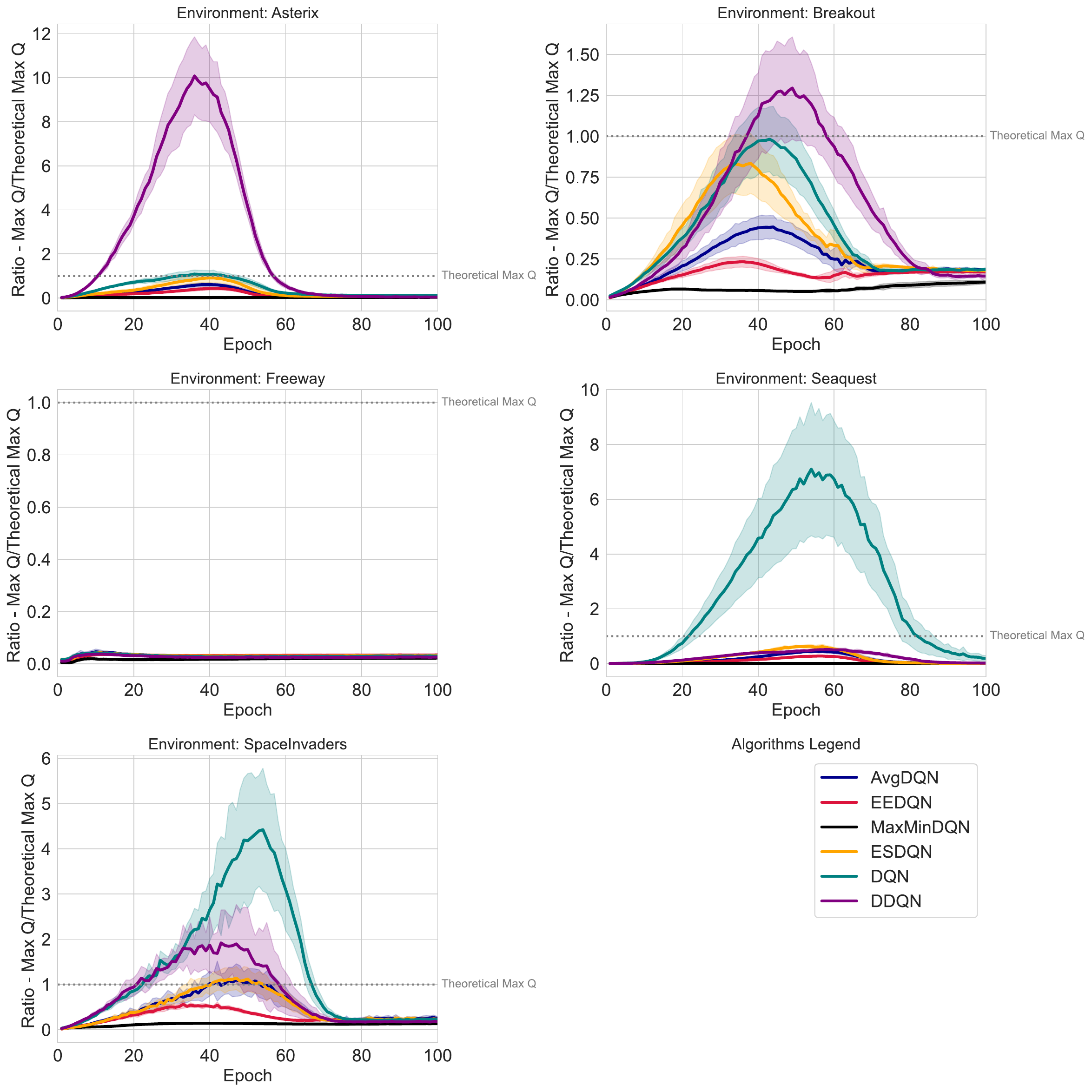}
\caption{Maximum observed Q-values normalised by the theoretical discounted return upper bound across training. The x-axis shows 100 training epochs, where each epoch corresponds to approximately 10,000 environment steps. The y-axis shows the maximum observed Q-value divided by $Q_{\max}=\frac{r_{\max}}{1-\gamma}$, so the dotted horizontal line at 1 marks the theoretical upper bound under the assumed reward limit. Curves show the mean across 10 seeds, and shaded regions denote the 95 percent confidence interval. Values above 1 indicate extreme value inflation.}
\label{fig2:ratioQ_minatar}
\end{figure*}

\subsection{Results and discussion}

Table~\ref{tab:eedqn_final_mean_reward} summarises final performance, measured as the mean episodic reward over the last 100 training episodes. EEDQN achieves the highest final return in four of the five environments: SpaceInvaders, Asterix, Breakout, and Freeway. In Seaquest, MaxMin DQN achieves the highest final return, while EEDQN remains the second-best method among the algorithms compared. The paired permutation tests show that EEDQN significantly outperforms several baselines across most environments, although not all pairwise comparisons are significant. In particular, the differences against AvgDQN in Asterix and ESDQN in Breakout are not statistically significant at the 0.05 level. Figure~\ref{fig1:average_minatar} further shows that EEDQN generally reaches its final performance plateau earlier, or at a comparable rate, in the environments where it achieves the strongest final performance.

Among the five benchmark environments, Freeway and Seaquest present the biggest exploration challenges, making them ideal testbeds for sparse reward evaluation. In Freeway, the agent must cross a freeway with moving cars at randomised speeds, yet receives reward only upon a successful crossing, with no intermediate feedback and also immediately resets after. Seaquest compounds difficulty through requiring agents to complete multiple objectives. The first objective is when enemies are destroyed which yields +1 reward, the second objective is to rescue divers and safely resurface which gives 10 reward. The strong empirical performance of both MaxMin DQN and Ensemble Elastic DQN (EEDQN) in these domains suggests that adaptive multi-step updates and conservative ensemble estimates can improve learning under sparse and delayed reward environments.

Figure~\ref{fig2:ratioQ_minatar} provides a diagnostic of extreme value inflation by plotting the maximum observed Q-value in each epoch divided by the theoretical discounted return upper bound. Values above the dotted horizontal line indicate Q-values that exceed the maximum possible discounted return under the assumed reward bound. EEDQN and MaxMin DQN remain below this threshold in all five environments, while DQN, DDQN, AvgDQN, and ESDQN exceed it in at least one environment. This suggests that EEDQN better controls extreme value inflation than the non-conservative baselines.

\begin{table*}[h]
\centering

\caption{Final performance comparison between EEDQN and benchmark algorithms, measured as the mean episodic reward over the last 100 training episodes. Bolded values indicate the best-performing algorithm for each environment. Reported p-values compare EEDQN against the algorithm in the corresponding row using a two-sided paired permutation test over 10 matched seeds. The EEDQN row has no p-value because it is the reference method.}
\begin{tabular}{@{}lllllllllll@{}}
\toprule
          & \multicolumn{2}{l}{Seaquest}     & \multicolumn{2}{l}{SpaceInvaders} & \multicolumn{2}{l}{Asterix}      & \multicolumn{2}{l}{Breakout}     & \multicolumn{2}{l}{Freeway}     \\ \midrule
          & mean           & p-val        & mean            & p-val        & mean           & p-val        & mean           & p-val        & mean          & p-val        \\ \midrule
AvgDQN    & 3.01           & \textless{}0.01 & 68.55           & \textless{}0.01 & 17.73          & 0.21            & 19.54          & \textless{}0.01 & 28.6          & \textless{}0.01 \\
MaxMinDQN & \textbf{27.58} & \textless{}0.01 & 90.18           & 0.05            & 3.18           & \textless{}0.01 & 13.54          & \textless{}0.01 & 39.1          & 0.05            \\
ESDQN     & 3.18           & \textless{}0.01 & 68.21           & \textless{}0.01 & 13.69          & \textless{}0.01 & 23.86          & 0.26            & 28.26         & \textless{}0.03 \\
DQN       & 0.65           & \textless{}0.01 & 33.96           & \textless{}0.01 & 4.76           & \textless{}0.01 & 13.81          & \textless{}0.01 & 20.03         & \textless{}0.04 \\
DDQN      & 0.68           & \textless{}0.01 & 24.71           & \textless{}0.01 & 1.95           & \textless{}0.01 & 10.35          & \textless{}0.01 & 14.14         & \textless{}0.05 \\
EEDQN     & 10.78          &                 & \textbf{101.42} &                 & \textbf{19.45} &                 & \textbf{25.77} &                 & \textbf{41.3} &                 \\ \bottomrule
\end{tabular}

\label{tab:eedqn_final_mean_reward}
\end{table*}

\subsection{EEDQN ablation study}
\label{sec:ablation}

We examined seven aggregation configurations to assess how ensemble predictions should be combined within the elastic step framework. 

\begin{itemize}
  \item \textbf{EEDQN}: uses the ensemble mean for one-step transitions and the ensemble minimum for multi-step transitions.
  \item \textbf{VariantEEDQN}: uses the ensemble minimum for one-step transitions and the ensemble mean for multi-step transitions.
  \item \textbf{MinEEDQN}: uses the ensemble minimum for all bootstrap targets.
  \item \textbf{MeanEEDQN}: uses the ensemble mean for all bootstrap targets.
  \item \textbf{ConvexEEDQN1}: uses a convex bootstrap value with $0.75$ weight on the ensemble mean and $0.25$ weight on the ensemble minimum.
  \item \textbf{ConvexEEDQN2}: uses a convex bootstrap value with $0.50$ weight on the ensemble mean and $0.50$ weight on the ensemble minimum.
  \item \textbf{ConvexEEDQN3}: uses a convex bootstrap value with $0.25$ weight on the ensemble mean and $0.75$ weight on the ensemble minimum.
\end{itemize}

For the convex variants, the bootstrap value is
\[
G_{\alpha}(s',a^{*}) =
\alpha \left(\frac{1}{N}\sum_{i=1}^{N}\hat Q_i(s',a^{*})\right)
+
(1-\alpha)\left(\min_i \hat Q_i(s',a^{*})\right),
\]
where $\alpha \in \{0.75,0.50,0.25\}$.

Table~\ref{tab:algorithm_ablation} shows that no single aggregation rule dominates across all environments. EEDQN achieves the highest final return in Asterix, Breakout, and Freeway, while ConvexEEDQN2 performs best in SpaceInvaders and MinEEDQN performs best in Seaquest. This pattern suggests that the appropriate degree of conservatism depends on the environment. In particular, the strong performance of MinEEDQN in Seaquest is consistent with the Q-value diagnostic in Figure~\ref{fig2:ratioQ_minatar}, where DQN exhibits especially severe value inflation and proceeds to underperform. Overall, the ablation indicates that EEDQN provides a favourable balance between mean-based and minimum-based aggregation, while more conservative aggregation may be advantageous in environments with stronger overestimation pressure. Because this ablation focuses on aggregation rules, it should be interpreted as evidence about how ensemble aggregation interacts with elastic return length rather than as a full component ablation of every part of the algorithm.

\begin{table*}[t]
\centering

\caption{Aggregation-rule ablation for EEDQN, measured as the average episodic reward over the last 100 training episodes. Bolded values indicate the highest-performing aggregation rule for each environment within this ablation sweep. The ESDQN row is included as a non-ensemble elastic-step reference point.}
\begin{tabular}{@{}llllll@{}}
\toprule
                                    & Seaquest       & SpaceInvaders   & Asterix        & Breakout       & Freeway       \\ \midrule
\multicolumn{1}{l|}{ESDQN}          & 3.18           & 68.21           & 13.69          & 23.86          & 28.26        \\
\multicolumn{1}{l|}{EEDQN}          & 10.78          & 101.42          & \textbf{19.45} & \textbf{25.77} & \textbf{41.3} \\
\multicolumn{1}{l|}{VariantEEDQN}   & 29.97          & 95.2            & 3.23           & 18.65          & 38.22         \\
\multicolumn{1}{l|}{MinEEDQN}       & \textbf{34.76} & 98.01           & 2.74           & 17.57          & 36.74         \\
\multicolumn{1}{l|}{MeanEEDQN}      & 9.02           & 100.07          & 17.3           & 25.19          & 38.79         \\
\multicolumn{1}{l|}{ConvexEEDQN1} & 11.6           & 102.99          & 19.3           & 24.74          & 38.93         \\
\multicolumn{1}{l|}{ConvexEEDQN2} & 12.22          & \textbf{105.88} & 14.85          & 23.81          & 37.47         \\
\multicolumn{1}{l|}{ConvexEEDQN3} & 9.29           & 100.52          & 5.38           & 20.31          & 36.01         \\ \bottomrule
\end{tabular}

\label{tab:algorithm_ablation}
\end{table*} 

\section{Conclusions and future work}

This paper introduced Ensemble Elastic DQN (EEDQN), a value-based reinforcement learning algorithm that combines adaptive elastic multi-step returns with ensemble-based target aggregation. The method replaces the clustering component of Elastic Step DQN with a lightweight Q-value difference rule, making elastic return construction simpler to apply in MinAtar-scale environments. Empirically, EEDQN improved over standard DQN and Elastic Step DQN on most tested environments and controlled the severe value inflation observed in several baselines under the proposed Q-value diagnostic. The aggregation rule ablation showed that the choice of ensemble aggregation matters, EEDQN achieved the strongest performance in three of the five environments, while more conservative variants performed better in Seaquest and a convex aggregation variant performed best in SpaceInvaders.

These results suggest that adaptive multi-step returns and ensemble aggregation can be complementary, but they also show that the best degree of conservatism is environment dependent. Future work should investigate adaptive aggregation rules that adjust the mean minimum trade-off online, larger ensemble sizes, and extensions to actor-critic methods where overestimation bias also affects policy improvement. A broader evaluation on larger benchmarks such as the Atari Learning Environment would further clarify how well the proposed mechanism scales beyond MinAtar.

\bibliographystyle{alpha}
\bibliography{sample}

@article{mnih2015human,
  title={Human-level control through deep reinforcement learning},
  author={Mnih, Volodymyr and Kavukcuoglu, Koray and Silver, David and Rusu, Andrei A and Veness, Joel and Bellemare, Marc G and Graves, Alex and Riedmiller, Martin and Fidjeland, Andreas K and Ostrovski, Georg and others},
  journal={nature},
  volume={518},
  number={7540},
  pages={529--533},
  year={2015},
  publisher={Nature Publishing Group UK London}
}

@article{obando2020revisiting,
  title={Revisiting rainbow: Promoting more insightful and inclusive deep reinforcement learning research},
  author={Obando-Ceron, Johan S and Castro, Pablo Samuel},
  journal={arXiv preprint arXiv:2011.14826},
  year={2020}
}

@article{young2019minatar,
  title={Minatar: An atari-inspired testbed for thorough and reproducible reinforcement learning experiments},
  author={Young, Kenny and Tian, Tian},
  journal={arXiv preprint arXiv:1903.03176},
  year={2019}
}

@article{van2018deep,
  title={Deep reinforcement learning and the deadly triad},
  author={Van Hasselt, Hado and Doron, Yotam and Strub, Florian and Hessel, Matteo and Sonnerat, Nicolas and Modayil, Joseph},
  journal={arXiv preprint arXiv:1812.02648},
  year={2018}
}

@article{lan2020maxmin,
  title={Maxmin Q-learning: Controlling the estimation bias of Q-learning},
  author={Lan, Qingfeng and Pan, Yangchen and Fyshe, Alona and White, Martha},
  journal={arXiv preprint arXiv:2002.06487},
  year={2020}
}

@inproceedings{anschel2017averaged,
  title={Averaged-DQN: Variance reduction and stabilization for deep reinforcement learning},
  author={Anschel, Oron and Baram, Nir and Shimkin, Nahum},
  booktitle={International conference on machine learning},
  pages={176--185},
  year={2017},
  organization={PMLR}
}

@article{d2021mushroomrl,
  title={Mushroomrl: Simplifying reinforcement learning research},
  author={D'Eramo, Carlo and Tateo, Davide and Bonarini, Andrea and Restelli, Marcello and Peters, Jan},
  journal={Journal of Machine Learning Research},
  volume={22},
  number={131},
  pages={1--5},
  year={2021}
}

@article{chen2022ensemble,
  title={Ensemble Reinforcement Learning in Continuous Spaces--A Hierarchical Multi-Step Approach for Policy Training},
  author={Chen, Gang and Huang, Victoria},
  journal={arXiv preprint arXiv:2209.14488},
  year={2022}
}

@inproceedings{van2016deep,
  title={Deep reinforcement learning with double Q-learning},
  author={Van Hasselt, Hado and Guez, Arthur and Silver, David},
  booktitle={Proceedings of the AAAI conference on artificial intelligence},
  volume={30},
  number={1},
  year={2016}
}

@article{nauman2024overestimation,
  title={Overestimation, overfitting, and plasticity in actor-critic: the bitter lesson of reinforcement learning},
  author={Nauman, Michal and Bortkiewicz, Micha{\l} and Mi{\l}o{\'s}, Piotr and Trzci{\'n}ski, Tomasz and Ostaszewski, Mateusz and Cygan, Marek},
  journal={arXiv preprint arXiv:2403.00514},
  year={2024}
}

@article{zhang2025beta,
  title={beta-DQN: Improving Deep Q-learning By Evolving the Behavior},
  author={Zhang, Hongming and Bai, Fengshuo and Xiao, Chenjun and Gao, Chao and Xu, Bo and M{\"u}ller, Martin},
  journal={arXiv preprint arXiv:2501.00913},
  year={2025}
}

@inproceedings{ly2023elastic,
  title={Elastic step DDPG: Multi-step reinforcement learning for improved sample efficiency},
  author={Ly, Adrian and Dazeley, Richard and Vamplew, Peter and Cruz, Francisco and Aryal, Sunil},
  booktitle={2023 International Joint Conference on Neural Networks (IJCNN)},
  pages={01--06},
  year={2023},
  organization={IEEE}
}

@inproceedings{he2020wd3,
  title={Wd3: Taming the estimation bias in deep reinforcement learning},
  author={He, Qiang and Hou, Xinwen},
  booktitle={2020 IEEE 32nd international conference on tools with artificial intelligence (ICTAI)},
  pages={391--398},
  year={2020},
  organization={IEEE}
}

@inproceedings{fujimoto2018addressing,
  title={Addressing function approximation error in actor-critic methods},
  author={Fujimoto, Scott and Hoof, Herke and Meger, David},
  booktitle={International conference on machine learning},
  pages={1587--1596},
  year={2018},
  organization={PMLR}
}

@inproceedings{thrun2014issues,
  title={Issues in using function approximation for reinforcement learning},
  author={Thrun, Sebastian and Schwartz, Anton},
  booktitle={Proceedings of the 1993 connectionist models summer school},
  pages={255--263},
  year={2014},
  organization={Psychology Press}
}

@article{ly2024elastic,
  title={Elastic step DQN: A novel multi-step algorithm to alleviate overestimation in Deep Q-Networks},
  author={Ly, Adrian and Dazeley, Richard and Vamplew, Peter and Cruz, Francisco and Aryal, Sunil},
  journal={Neurocomputing},
  volume={576},
  pages={127170},
  year={2024},
  publisher={Elsevier}
}

@inproceedings{hessel2018rainbow,
  title={Rainbow: Combining improvements in deep reinforcement learning},
  author={Hessel, Matteo and Modayil, Joseph and Van Hasselt, Hado and Schaul, Tom and Ostrovski, Georg and Dabney, Will and Horgan, Dan and Piot, Bilal and Azar, Mohammad and Silver, David},
  booktitle={Proceedings of the AAAI conference on artificial intelligence},
  volume={32},
  number={1},
  year={2018}
}

@article{yuan2019novel,
  title={A novel multi-step Q-learning method to improve data efficiency for deep reinforcement learning},
  author={Yuan, Yinlong and Yu, Zhu Liang and Gu, Zhenghui and Yeboah, Yao and Wei, Wu and Deng, Xiaoyan and Li, Jingcong and Li, Yuanqing},
  journal={Knowledge-Based Systems},
  volume={175},
  pages={107--117},
  year={2019},
  publisher={Elsevier}
}

@article{hernandez2019understanding,
  title={Understanding multi-step deep reinforcement learning: A systematic study of the DQN target},
  author={Hernandez-Garcia, J Fernando and Sutton, Richard S},
  journal={arXiv preprint arXiv:1901.07510},
  year={2019}
}

\appendix
\section{Hyper-parameters}
\label{apd:first}

\begin{table}[H]
\centering
\scriptsize
\renewcommand{\arraystretch}{0.92}
\caption{Hyper-parameters used across the MinAtar environments. All algorithms used the same hyper-parameters unless otherwise specified.}
\label{tab:hyperparameters}
\begin{tabular}{@{}p{0.30\textwidth}p{0.64\textwidth}@{}}
\toprule
\textbf{Hyper-parameter} & \textbf{Value} \\
\midrule
\multicolumn{2}{@{}l}{\textbf{Training setup}} \\
\midrule
Environment suite & MinAtar: Seaquest, SpaceInvaders, Asterix, Breakout, and Freeway \\
Training budget & 1,000,000 environment steps per seed \\
Random seeds & 10 matched seeds per algorithm \\
Discount factor $\gamma$ & 0.99 \\
Batch size & 32 \\
Replay memory size & 100,000 transitions \\
Initial replay size & 5,000 random transitions \\
Training frequency & One gradient update per environment step after replay initialisation \\
Target update interval & 1000 environment steps \\
Final evaluation metric & Mean episodic return over the last 100 training episodes \\
\midrule
\multicolumn{2}{@{}l}{\textbf{Optimisation and exploration}} \\
\midrule
Optimiser & Adam \\
Learning rate & $2.5\times10^{-4}$ \\
Loss function & Mean squared TD error \\
Exploration policy & $\epsilon$-greedy \\
Exploration schedule & Linear decay from $\epsilon=1.0$ to $\epsilon=0.1$ \\
Exploration decay length & 1,000,000 environment steps \\
Evaluation $\epsilon$ & 0.05 \\
\midrule
\multicolumn{2}{@{}l}{\textbf{Shared network architecture}} \\
\midrule
Network type & Convolutional Q-network used for all algorithms \\
Convolutional layer & 16 filters, kernel size $3\times3$, stride 1 \\
Convolutional activation & ReLU \\
Flattening & Flatten convolutional feature map before the fully connected layers \\
Hidden fully connected layer & 128 hidden units \\
Hidden activation & ReLU \\
Output layer & Linear layer with one output per discrete action \\
Weight initialisation & Xavier uniform initialisation; ReLU gain for hidden layers and linear gain for the output layer \\
\midrule
\multicolumn{2}{@{}l}{\textbf{Algorithm-specific settings}} \\
\midrule
Backup horizon & Fixed one-step for DQN, DDQN, AvgDQN, and MaxMinDQN; adaptive for ESDQN and EEDQN \\
Ensemble size & 2 for AvgDQN, MaxMinDQN, and EEDQN \\
Ensemble member initialisation & Independent random initialisation for each ensemble member \\
EEDQN aggregation rule & Ensemble mean for one-step targets; ensemble minimum for longer elastic returns \\
Value-difference buffer size & 10,000 for EEDQN \\
Adaptive threshold & $\operatorname{mean}(B)+\operatorname{std}(B)/\sqrt{|B|}$ \\
\bottomrule
\end{tabular}
\end{table}

\end{document}